\documentclass{article}
\usepackage{spconf,amsmath,graphicx}
\usepackage{booktabs,multirow}
\usepackage{amsmath}
\usepackage{xcolor}
\usepackage{soul,color}
\usepackage{enumitem}
\setlist{nosep, leftmargin=14pt}

\usepackage{mwe} 

\definecolor{pose_auto}{HTML}{ff0000}
\definecolor{pose_gt}{HTML}{008402}
\definecolor{bbox}{HTML}{e1e1e1}

\title{Predicting Coronary Artery Calcium Severity based on Non-Contrast Cardiac CT images using Deep Learning}
\name{Lachlan Nguyen$^{\star}$,
Aidan Cousins$^{\dagger}$,
Arcot Sowmya$^{\ddagger}$
Hugh Dixson$^{\star, \S}$,
Sonit Singh $^{\ddagger}$}
\address{$^{\star}$ School of Clinical Medicine, Faculty of Medicine \& Health, UNSW Sydney, Australia \\
$^{\dagger}$ Rural Clinical School, Faculty of Medicine \& Health, UNSW, Port Macquarie, Australia\\
$^{\ddagger}$ School of Computer Science and Engineering, Faculty of Engineering, UNSW Sydney, Australia \\
$^{\S}$Bankstown-Lidcombe Hospital, South Western Sydney Local Health District, Bankstown, Australia}

\begin{document}
%
\maketitle
\begin{abstract}
Cardiovascular disease causes high rates of mortality worldwide. Coronary artery calcium (CAC) scoring is a powerful tool to stratify the risk of atherosclerotic cardiovascular disease. Current scoring practices require time-intensive semiautomatic analysis of cardiac computed tomography by radiologists and trained radiographers. The purpose of this study is to develop a deep learning convolutional neural networks (CNN) model to classify the calcium score in cardiac, non-contrast computed tomography images into one of six clinical categories. A total of 68 patient scans were retrospectively obtained together with their respective reported semiautomatic calcium score using an ECG-gated GE Discovery 570 Cardiac SPECT/CT camera. The dataset was divided into training, validation and test sets. Using the semiautomatic CAC score as the reference label, the model demonstrated high performance on a six-class CAC scoring categorisation task. Of the scans analysed, the model misclassified 32 cases, tending towards overestimating the CAC in 26 out of 32 misclassifications. Overall, the model showed high agreement (Cohen’s $kappa$ = 0.962), an overall accuracy of $96.5\%$ and high generalisability. The results suggest that the model outputswere accurate and consistent with current semiautomatic practice, with good generalisability to test data. The model demonstrates the viability of a CNN model to stratify the calcium score into an expanded set of six clinical categories. 
\end{abstract}

\section{Introduction}~\label{sec:intro}

Cardiovascular disease (CVD) is a leading cause of morbidity and mortality worldwide, and is the most common cause of Australian deaths ($31\%$ of all deaths)~\cite{AIHW_CVD}. At the forefront of CVD is coronary artery disease representing $41\%$ of cardiovascular deaths, most commonly due to atherosclerotic plaque formation~\cite{AIHW_Heart_Stroke}. The natural history of plaque formation is the progressive calcification of the atherosclerotic plaque~\cite{Kalampogias:2016}. These coronary artery calcifications can be assessed to identify patients at risk for cardiovascular events using computed tomography (CT)~\cite{Hata:2022:Aortic_calcification}.

Coronary artery calcium (CAC) scoring involves quantifying the calcium within coronary arteries through an electrocardiogram (ECG) gated, non-contrast CT scan of the heart~\cite{Hamilton_Craig:2017:CSA}. Scan acquisition is fast, exposes the patient to minimal radiation, and does not require any intravenous contrast~\cite{Hamilton_Craig:2017:CSA}. It involves multidetector (at least 64-slice) scanners with fast gantry rotation speeds (420 ms or shorter) and good temporal / spatial resolution, producing accurate and reliable coronary artery visualisation. Scanners are ECG-linked to synchronise the image acquisition to the cardiac cycle. Current clinical practice involves manual assessment of the CT scan by a radiographer or radiologist to segment the arteries and highlight calcified areas. This is fed into semi-automatic software to produce a quantified Agatston score based on a threshold of 130 Hounsfield units~\cite{Hong:2022:automated_coronary}. Whilst this process is widely used in research and clinical contexts it does have limitations, including varied observer scoring with a reported inter-reader variability of $3\%$, an intra-reader variability of $<1\%$, and an interscan variability of $15\%$~\cite{George:2008:Calcium_scores}. The manual process is also time consuming and labour intensive,  therefore an improved and efficient stratifying technique would be highly useful~\cite{Hong:2022:automated_coronary}. 

There have been many deep learning models utilising convolutional neural networks (CNNs) to automate CAC scoring. Key studies have found that fully automated models demonstrated high accuracy and excellent agreement with  reference standards with low analysis times~\cite{Eng:2021:Automated_calcium, Ihdayhid:2023:evaluation}. However, most studies involved a large proportion of zero CAC data, with some only considering total scoring without considering individual artery calcium~\cite{Ihdayhid:2023:evaluation}. Another study, whilst highly accurate,  tended to overestimate when misclassifying CAC~\cite{Takahashi:2023}. It was observed that misclassifications were more common within lower calcific categories (from a CAC score of 1-100)~\cite{Martin:2020, Takahashi:2023}. It is clear that current studies produce high accuracy with close agreement with reference standards. These, however, are often based on a five class risk categorisation ranging from $0$: No evidence of CAD, $1-10$: Minimal, $11-100$: Mild, $101-400$: Moderate, and $>400$: Severe. This five-class risk framework may oversimplify patient risk stratification,  whereas  a six class framework could allow for finer and more nuanced risk differentiation particularly in the moderate to high risk categories. 

The aim of this study was to train and assess the effectiveness of a deep learning CNN model to predict CAC scoring in a six clinical category framework, based on cardiac, non-contrast ECG-gated CT image analysis. The six classes include: 0:  Zero, 0-10: Low, 10-100: Low/ Moderate, 101-400: Moderate, 401-1000: Moderate/ High, and $>$1000: High. The model is hypothesised to demonstrate similar accuracy compared to current practice, with potential benefits in efficiency and decreased labour intensity. An accurate and successful model would significantly impact clinical practice through faster risk stratification, by reducing the workload on radiographers / radiologists and reducing interobserver variability.

\section{Materials and Method}~\label{sec:method}
\begin{figure}
    \centering
    \includegraphics[scale=0.64]{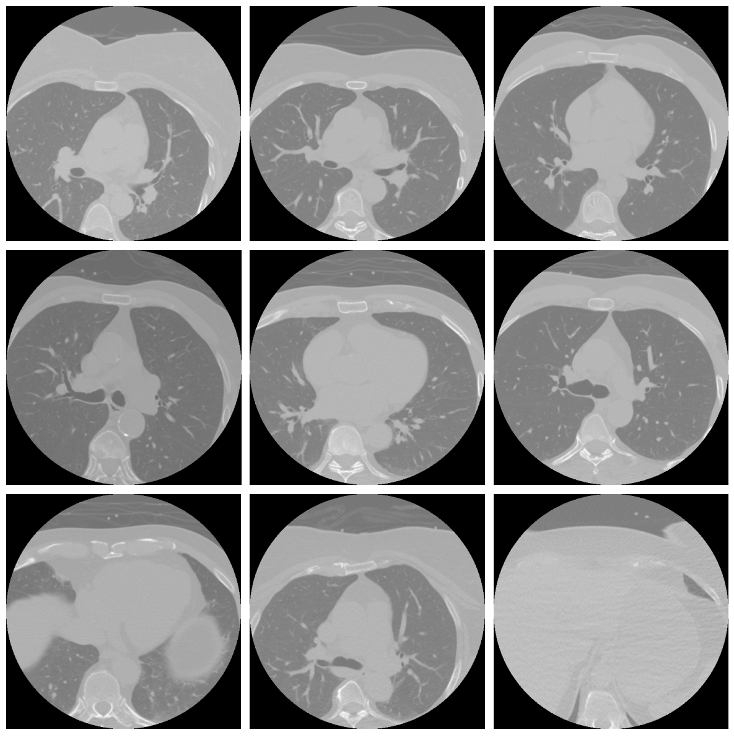}
    \caption{Sample Non-contrast cardiac CT images used in this study.}
    \label{fig:sample_images}
\end{figure}

This was a retrospective study conducted remotely and approved by the South Western Sydney Local Health District Human Research Ethics Committee (ETH01335). Data collection was done between the period Feb 22-June 13 2024. The scans were acquired using the CT component of a GE Discovery 570 Cardiac SPECT/CT camera, which is equivalent to a GE Lightspeed VCT 64 machine, and all scans were ECG-gated. All patients who obtained a calcium score within this period at the Bankstown-Lidcombe Hospital were included in the dataset, with each consecutive scan given a consecutive study number, and their reported calcium score entered into a spreadsheet. This spreadsheet, containing the study number and its calcium score, determined the ground truth labels. All patient data was removed from the DICOM files, with each scan identified only by its study number. This led to the collection of 68 patients' scans, with 56 to 96 slices per study. The full dataset was split into test (30\%), validation (21\%)  and training (49\%) datasets. Examples of images can be seen in Figure~\ref{fig:sample_images}.

\begin{figure}
    \centering
    \includegraphics[scale=0.55]{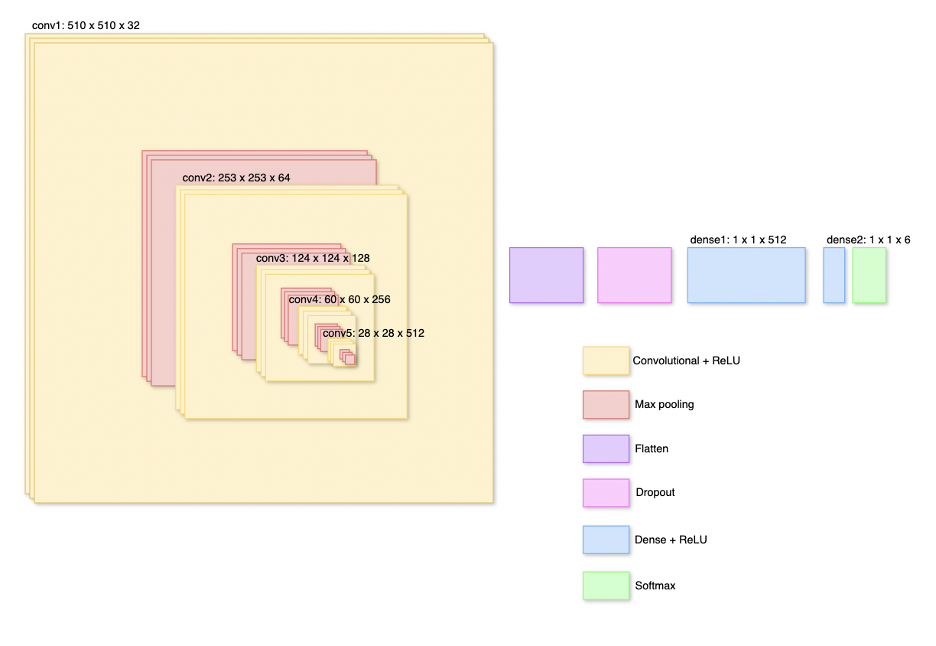}
    \caption{Customised CNN architecture for predicting severity of calcium in CTCA images.}
    \label{fig:model}
\end{figure}

We built a customised Convolutional Neural Networks (CNN) for the classification task. The network architecture of the proposed model is in Figure~\ref{fig:model}. The network consisted of five convolutional layers, where the first layer applied 32 filters and subsequently doubled the number of  filters in every convolutional layer, reaching a maximum of 512 filters. Each convolutional layer applied a kernel size of (3, 3) and a ReLU activation function, followed by a max-pooling layer of size (2, 2). The output of the convolutions and max-pooling was flattened into a 1-dimensional vector with a 30\% dropout applied. The network had two  fully connected dense layers at the end. The first consisted of 512 neurons and a ReLU activation function, and the final layer consisted of 6 output neurons (representing the 6 calcium categories) and a softmax activation function. The network architecture was refined and optimised by trialling different configurations of the layers, dropout and neurons and comparing the performance. 

\section{Results}~\label{sec:results}

All the experiments were implemented in Python version 3.10.8 using PyTorch~\cite{PyTorch}. The experimental setup involved compiling the model with an \emph{Adam} optimiser~\cite{Adam}, initially set to a learning rate of $1e^-4$, a \emph{categorical cross-entropy} loss and an \emph{accuracy} metric. The class imbalance was addressed by calculating class weights to emphasise underrepresented classes during training. Additionally, a learning rate reduction strategy was implemented to monitor validation loss, halving the learning rate if it does not improve in two epochs. The reduction was allowed until a minimum threshold of $1e^-6$ was reached. The model was trained and validated on the training and validation data in $15$ epochs with a batch size of $32$. Results were obtained  by five fold cross-validation, ensuring patient mutual exclusivity. We report the results of model testing in terms of classification metrics, namely precision, recall and F1-score. In this context, precision refers to the proportion of classifications which were assigned the correct class. Recall refers to the proportion of classes which were correctly identified, with the F1-score representing the harmonic mean of  precision and recall, providing a balanced metric of model accuracy~\cite{Hicks:2022:On_evaluation_metrics}. 

The results of model performance in terms of the classification metrics are given in Table~\ref{tab:results}. The results demonstrate the model's ability to distinguish scans with very little to very high calcification quite accurately. Overall, the model performed very well, with high accuracy and strong agreement with the ground truth scores (Cohen's $\kappa$ = 0.962). 

\begin{table}[]
    \centering
    \caption{Performance of the proposed model in terms of classification metrics for predicting severity of the calcium in CT images.}
    \label{tab:results}
    \begin{tabular}{lccc}
    \toprule
       \textbf{Category}  & \textbf{Precision} & \textbf{Recall} & \textbf{F1-score}  \\
    \midrule
    Very Low Risk & 0.998 & 0.976 & 0.986 \\
    Low Risk & 0.970 & 0.996 & 0.984 \\
    Low/Moderate Risk & 0.972 & 0.952 & 0.964 \\
    Moderate Risk & 0.932 & 0.928 & 0.932 \\
    Moderate/High Risk & 0.970 & 0.990 & 0.978 \\
    High Risk & 0.948 & 0.952 & 0.948 \\
    \midrule
    \textbf{Average} & \textbf{0.965} & \textbf{0.965} & \textbf{0.965} \\
    \bottomrule
    \end{tabular}
\end{table}

\begin{figure}
    \centering
    \includegraphics[width=\linewidth]{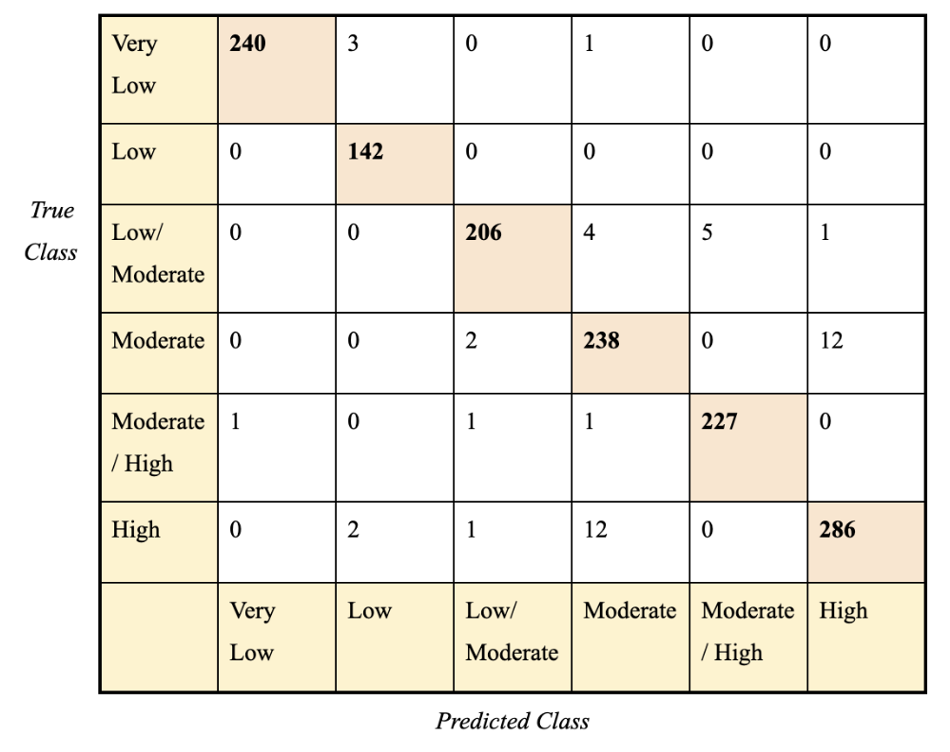}
    \caption{Confusion matrix showing true vs predicted labels for 6-class classification.}
    \label{fig:confusion_matrix}
\end{figure}

To further analyse the model’s performance on individual class predictions, a \emph{confusion matrix} was produced, see  Figure~\ref{fig:confusion_matrix}. The confusion matrix shows  high true positive rates for each class with minimal misclassification. “Very Low” class predictions were 98.4\% correct (240 out of 244 cases correctly classified). The misclassifications were minor, identifying some cases as “Low” (3) and “Moderate” (1). “Low” predictions were completely correctly classified  (142 cases) yielding 100\% accuracy. “Low/ Moderate” accurately classified 206 samples (out of 216, 95.4\% accuracy) with misclassification tending towards predicting higher calcific categories. Similarly, “Moderate” classification had an accuracy of 94.4\% (238 out of 252 cases), misclassifying 12 cases as “High” and 2 as “Low/ Moderate”. “Moderate/ High” classifications were highly accurate (98.7\%, 227 of 230 samples classified correctly) with minor misclassification tending towards predicting lower calcific categories. Finally, “High” classification showed 95\% accuracy (286 out of 301 samples). Overall, the model produced high accuracy, with most misclassifications tending towards predicting higher calcific categories (26 out of 32). 

\begin{figure}
    \centering
    \includegraphics[scale=0.58]{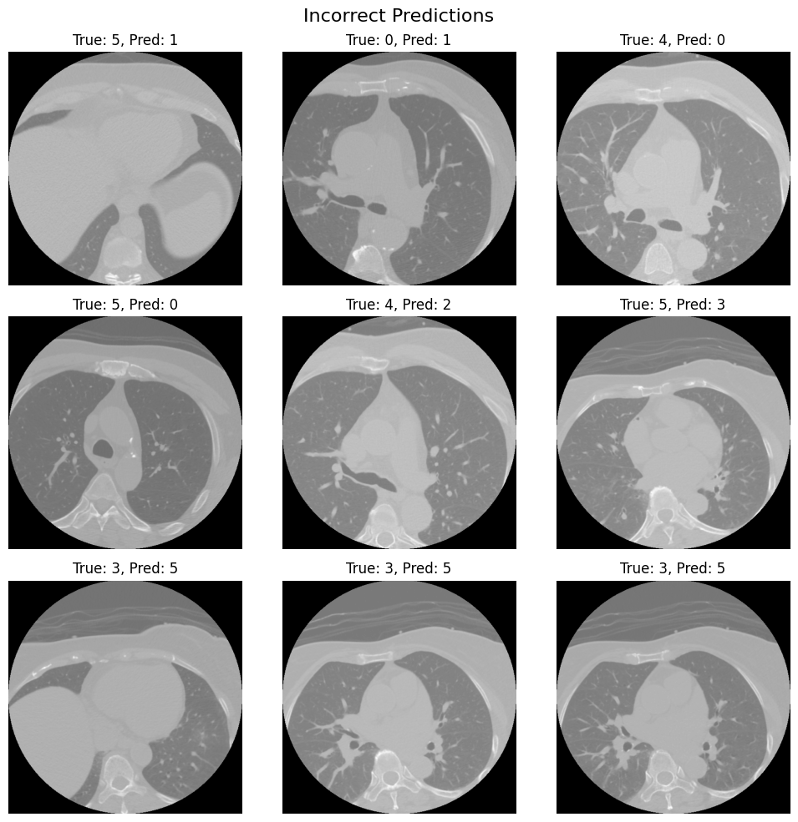}
    \caption{Qualitative results showing misclassified scans with their true and predicted class label.}
    \label{fig:qualitative_results}
\end{figure}

Assessing the images themselves alongside the true and predicted classes, as shown in Figure~\ref{fig:qualitative_results}, illustrates that the model’s misclassifications are often due to slice based analysis. The model accounts for the patient’s total calcium score, which may not be present within each image slice. For example, the patient's true CAC score may be “High” however the image slice of the heart apex tip is unlikely to show high calcium thus prompting a low calcium prediction. Similarly, the model tends to select incorrect features with some artefact-driven classifications. Features including the thoracic ribs and vertebrae commonly seen in cardiac CT scans possibly bias the model’s classifications towards higher calcific categories, hence accounting for this model’s tendency to misclassify scans into higher CAC categories. 

\section{Discussion}~\label{sec:discussion}
The aim of this study was to determine the success and accuracy of a CNN model to classify CAC score into six clinical categories in cardiac, non-contract CT images. The categorisations were highly accurate and were achieved by  obtaining relevant scans and their true calcium scores to train / test a customised CNN. CNN models are known for effectiveness in image analysis and are accurate when applied to cardiac image analysis for calcium scoring, and therefore were chosen for this study~\cite{Eng:2021:Automated_calcium, Hong:2022:automated_coronary, Ihdayhid:2023:evaluation, Martin:2020, Jamka:2024, Takahashi:2023, Wang:2020, Winkel:2021, Xu:2021}. Using current manual practices to obtain reference labels to train and assess model performance is common~\cite{Zeleznik:2021}, and this study adopted the same approach. Similar studies used metrics such as precision, recall, F1-score, and Cohen’s $\kappa$ to measure model performance~\cite{Hong:2022:automated_coronary}, and this was replicated in this study. Given the six-class output, a customised CNN was developed using an existing architecture, thereby providing greater task-specific design flexibility without complexities introduced by more generalised models~\cite{Litjens:2017:survey}. Furthermore, the design simplicity allowed for lower computational requirements, less overfitting and increased interpretability by clinicians~\cite{Rudin:2019}. 

The model was highly accurate and consistent across six clinical categories based on the performance metrics (precision, recall and F1-score average = 0.965), with close agreement with the reference standard (Cohen’s $kappa$ = 0.962). This aligns with current literature supporting the plausibility of predictive AI models within clinical predictive tasks for cardiovascular risk assessment. One study applied a modified CNN with ResNet architecture to perform automated calcium scoring, demonstrating no significant difference between predictions and their annotated reference with a 2.7 second processing time~\cite{Litjens:2017:survey}. This utilised the risk categories (0, 1 to 10, 11 to 100, 101 to 400, and $>$400) with accuracy for each class being 94.9\% (258 of 272), 73.5\% (36 of 49), 95.6\% (87 of 91), 98.3\% (59 of 60), and 92.3\% (36 of 39), respectively~\cite{Martin:2020}. Another study similarly found success using the five risk categories with a model analysis time of $13.1 \pm 3.2$ seconds/scan, a strong correlation between predicted scores and manual scoring (ICC = 0.98) and high agreement (Cohen’s $kappa$ = 0.90)~\cite{Ihdayhid:2023:evaluation}. Our study extends these findings in multi-class categorisation to six classes. In particular, “Zero” and “Low” classification are vital to identify early-stage calcifications and at-risk patients, which is achieved with a high precision and f1 score. Our findings show that CNNs applied to cardiac non-contrast CT scans perform better in classifying lower than higher calcific categories~\cite{Eng:2021:Automated_calcium}. This may be due to more stark differences between zero and minimal calcium compared to subtle differences between moderate and high. Whilst this is promising for distinguishing no risk patients from those at some risk, it may reflect an issue with CNN models when applied to high calcium volume scans. Analysing the images themselves, artefact bias tends to push classifications towards higher CAC categories. To address this, segmentation or ROI cropping in data pre-processing aimed at removing non-cardiac structures (e.g., thoracic ribs and vertebrae) will reduce artefact-driven misclassifications. The current literature often uses a five-class scoring system. This study shows the plausibility of increasing the number of classes, from five to six class categorisation, potentially providing a more nuanced classification system. This study demonstrated potential to accommodate for alterations to the CAC risk scoring system illustrating how CNN models can adapt without reductions in accuracy or agreement.

In terms of limitations, the model analysis was limited to 68 patients. A more balanced and broad dataset may better represent the model’s performance and generalisability, as this dataset contained fewer “Low” class scans. To improve performance in future model versions, data augmentation can be implemented, which is often used to address issues with bias, overfitting and inaccuracy within datasets. Furthermore, this study utilises the calcium score produced by the existing  semiautomatic software as the reference standard. Whilst this method is accurate, the gold standard for accuracy remains manual annotation and assessment by expert readers /  radiologists~\cite{Winkel:2021}. This study was based on data collected from a single site using a single scanner. In future work, the study could collate data over a larger time frame providing more data for model training / testing. Additionally, scans from different hospitals and scanners would be beneficial to build a more generalisable model and provide a better representation for real-world and widespread clinical use.

\section{Conclusion}~\label{sec:conclusion}
This study developed a customised CNN model capable of classifying the coronary vessel calcium score into six clinical categories illustrating flexibility to alterations in the current scoring system. The experimental results demonstrated that deep learning has potential to correctly categorise calcifications in CT images based on its severity. The proposed system could potentially be used in clinical settings to detect early signs of calcifications, saving human lives. 



\section{Compliance with ethical standards}~\label{sec:ethics}
This study was performed in line with the principles of the Declaration of Helsinki. The CTCA scans used for this work were obtained as part of an ethically approved study (reference number: 2024/ETH01335) performed at a tertiary referral hospital in Australia, in 2023-24.

\section{Acknowledgments}~\label{sec:acknowledgments}

No funding was received for conducting this study. The authors have no relevant financial or non-financial interests to disclose.

\bibliographystyle{IEEEbib}
\bibliography{refs}

@ARTICLE{Baebler:2023:AI_in_CTCA,
AUTHOR={Baeßler, Bettina  and Götz, Michael  and Antoniades, Charalambos  and Heidenreich, Julius F.  and Leiner, Tim  and Beer, Meinrad },       
TITLE={Artificial intelligence in coronary computed tomography angiography: Demands and solutions from a clinical perspective},      
JOURNAL={Frontiers in Cardiovascular Medicine},     
VOLUME={Volume 10 - 2023},
YEAR={2023},
URL={https://www.frontiersin.org/journals/cardiovascular-medicine/articles/10.3389/fcvm.2023.1120361},
DOI={10.3389/fcvm.2023.1120361},
ISSN={2297-055X},
}

@Article{Eng:2021:Automated_calcium,
author={Eng, David
and Chute, Christopher
and Khandwala, Nishith
and Rajpurkar, Pranav
and Long, Jin
and Shleifer, Sam
and Khalaf, Mohamed H.
and Sandhu, Alexander T.
and Rodriguez, Fatima
and Maron, David J.
and Seyyedi, Saeed
and Marin, Daniele
and Golub, Ilana
and Budoff, Matthew
and Kitamura, Felipe
and Takahashi, Marcelo Straus
and Filice, Ross W.
and Shah, Rajesh
and Mongan, John
and Kallianos, Kimberly
and Langlotz, Curtis P.
and Lungren, Matthew P.
and Ng, Andrew Y.
and Patel, Bhavik N.},
title={Automated coronary calcium scoring using deep learning with multicenter external validation},
journal={npj Digital Medicine},
year={2021},
month={Jun},
day={01},
volume={4},
number={1},
pages={88},
issn={2398-6352},
doi={10.1038/s41746-021-00460-1},
url={https://doi.org/10.1038/s41746-021-00460-1}
}

@article{George:2008:Calcium_scores,
doi = {10.2174/1874192400802010087},
author = {George Anil and Movahed Assad and },
title = {Coronary Artery Calcium Scores: Current Thinking and Clinical Applications},
journal = {The Open Cardiovascular Medicine Journal},
volume = {2},
year = {2008/9/18},
doi = {10.2174/1874192400802010087},
URL = {https://opencardiovascularmedicinejournal.com/abs/10.2174/1874192400802010087},
}

@article{Hamilton_Craig:2017:CSA,
author = {Hamilton-Craig, Christian R and Chow, Clara K and Younger, John F and Jelinek, V M and Chan, Jonathan and Liew, Gary YH},
title = {Cardiac Society of Australia and New Zealand position statement executive summary: coronary artery calcium scoring},
journal = {Medical Journal of Australia},
volume = {207},
number = {8},
pages = {357-361},
doi = {https://doi.org/10.5694/mja16.01134},
url = {https://onlinelibrary.wiley.com/doi/abs/10.5694/mja16.01134},
year = {2017}
}

@Article{Hata:2022:Aortic_calcification,
author={Hata, Yoshiki
and Mochizuki, Junji
and Okamoto, Shuichi
and Matsumi, Hiroaki
and Hashimoto, Katsushi},
title={Aortic calcification is associated with coronary artery calcification and is a potential surrogate marker for ischemic heart disease risk: A cross-sectional study},
journal={Medicine},
year={2022},
volume={101},
number={29},
url={https://journals.lww.com/md-journal/fulltext/2022/07220/aortic_calcification_is_associated_with_coronary.11.aspx}
}

@Article{Hicks:2022:On_evaluation_metrics,
author={Hicks, Steven A.
and Str{\"u}mke, Inga
and Thambawita, Vajira
and Hammou, Malek
and Riegler, Michael A.
and Halvorsen, P{\aa}l
and Parasa, Sravanthi},
title={On evaluation metrics for medical applications of artificial intelligence},
journal={Scientific Reports},
year={2022},
month={Apr},
day={08},
volume={12},
number={1},
pages={5979},
issn={2045-2322},
doi={10.1038/s41598-022-09954-8},
url={https://doi.org/10.1038/s41598-022-09954-8}
}

@article{Hong:2022:automated_coronary,
title = {Automated coronary artery calcium scoring using nested U-Net and focal loss},
journal = {Computational and Structural Biotechnology Journal},
volume = {20},
pages = {1681-1690},
year = {2022},
issn = {2001-0370},
doi = {https://doi.org/10.1016/j.csbj.2022.03.025},
url = {https://www.sciencedirect.com/science/article/pii/S2001037022001003},
author = {Jia-Sheng Hong and Yun-Hsuan Tzeng and Wei-Hsian Yin and Kuan-Ting Wu and Huan-Yu Hsu and Chia-Feng Lu and Ho-Ren Liu and Yu-Te Wu},
}

@Article{Ihdayhid:2023:evaluation,
author={Ihdayhid, Abdul Rahman
and Lan, Nick S. R.
and Williams, Michelle
and Newby, David
and Flack, Julien
and Kwok, Simon
and Joyner, Jack
and Gera, Sahil
and Dembo, Lawrence
and Adler, Brendan
and Ko, Brian
and Chow, Benjamin J. W.
and Dwivedi, Girish},
title={Evaluation of an artificial intelligence coronary artery calcium scoring model from computed tomography},
journal={European Radiology},
year={2023},
month={Jan},
day={01},
volume={33},
number={1},
pages={321-329},
issn={1432-1084},
doi={10.1007/s00330-022-09028-3},
url={https://doi.org/10.1007/s00330-022-09028-3}
}

@article{Kalampogias:2016,
 doi = {10.2174/1573406411666150928111446},
 Title = {Basic Mechanisms in Atherosclerosis: The Role of Calcium},
 Journal = {Medicinal Chemistry},
 Volume = {12},
 Number = {2},
 Pages = {103-113},
 Year = {2016},
 ISSN = {1573-4064/1875-6638},
 DOI = {10.2174/1573406411666150928111446},
 URL = {https://www.eurekaselect.com/article/70643},
 Author = {Aimilios Kalampogias},
 Author = {Gerasimos Siasos},
 Author = {Evangelos Oikonomou},
 Author = {Sotirios Tsalamandris},
 Author = {Konstantinos Mourouzis},
 Author = {Vasiliki Tsigkou},
 Author = {Manolis Vavuranakis},
 Author = {Thodoris Zografos},
 Author = {Spyridon Deftereos},
 Author = {Christodoulos Stefanadis},
 Author = {Dimitris Tousoulis},
 }

@ARTICLE{Liao:2022:AI_in_CTCA,
AUTHOR={Liao, Jiahui  and Huang, Lanfang  and Qu, Meizi  and Chen, Binghui  and Wang, Guojie },      
TITLE={Artificial Intelligence in Coronary CT Angiography: Current Status and Future Prospects},      
JOURNAL={Frontiers in Cardiovascular Medicine},      
VOLUME={Volume 9 - 2022},
YEAR={2022},
URL={https://www.frontiersin.org/journals/cardiovascular-medicine/articles/10.3389/fcvm.2022.896366},
DOI={10.3389/fcvm.2022.896366},
ISSN={2297-055X},
}

@article{Litjens:2017:survey,
title = {A survey on deep learning in medical image analysis},
journal = {Medical Image Analysis},
volume = {42},
pages = {60-88},
year = {2017},
issn = {1361-8415},
doi = {https://doi.org/10.1016/j.media.2017.07.005},
url = {https://www.sciencedirect.com/science/article/pii/S1361841517301135},
author = {Geert Litjens and Thijs Kooi and Babak Ehteshami Bejnordi and Arnaud Arindra Adiyoso Setio and Francesco Ciompi and Mohsen Ghafoorian and Jeroen A.W.M. {van der Laak} and Bram {van Ginneken} and Clara I. Sánchez},
}

@article{Martin:2020,
title = {Evaluation of a Deep Learning–Based Automated CT Coronary Artery Calcium Scoring Algorithm},
journal = {JACC: Cardiovascular Imaging},
volume = {13},
number = {2, Part 1},
pages = {524-526},
year = {2020},
issn = {1936-878X},
doi = {https://doi.org/10.1016/j.jcmg.2019.09.015},
url = {https://www.sciencedirect.com/science/article/pii/S1936878X19309386},
author = {Simon S. Martin and Marly {van Assen} and Saikiran Rapaka and H. Todd Hudson and Andreas M. Fischer and Akos Varga-Szemes and Pooyan Sahbaee and Chris Schwemmer and Mehmet A. Gulsun and Serkan Cimen and Puneet Sharma and Thomas J. Vogl and U. Joseph Schoepf}
}

@article{Jamka:2024,
    author = {Miszalski-Jamka, K and Malawski, F and Goslinski, J and Witkowska, M and Purgol, P and Bujny, M and Bartczak, T and Malara, W and Banach, M and Kostur, M},
    title = {{AI}-based algorithm for assessing coronary artery calcium score on contrast-enhanced cardiac computed tomography scans},
    journal = {European Heart Journal},
    volume = {45},
    number = {Supplement 1},
    pages = {ehae666.197},
    year = {2024},
    month = {10},
    issn = {0195-668X},
    doi = {10.1093/eurheartj/ehae666.197},
    url = {https://doi.org/10.1093/eurheartj/ehae666.197},
}

@Article{Rudin:2019,
author={Rudin, Cynthia},
title={Stop explaining black box machine learning models for high stakes decisions and use interpretable models instead},
journal={Nature Machine Intelligence},
year={2019},
month={May},
day={01},
volume={1},
number={5},
pages={206-215},
issn={2522-5839},
doi={10.1038/s42256-019-0048-x},
url={https://doi.org/10.1038/s42256-019-0048-x}
}

@article{Takahashi:2023,
    author = {Takahashi, Daigo and Fujimoto, Shinichiro and Nozaki, Yui O and Kudo, Ayako and Kawaguchi, Yuko O and Takamura, Kazuhisa and Hiki, Makoto and Sato, Eisuke and Tomizawa, Nobuo and Daida, Hiroyuki and Minamino, Tohru},
    title = {Fully automated coronary artery calcium quantification on electrocardiogram-gated non-contrast cardiac computed tomography using deep-learning with novel Heart-labelling method},
    journal = {European Heart Journal Open},
    volume = {3},
    number = {6},
    pages = {oead113},
    year = {2023},
    month = {11},
    issn = {2752-4191},
    doi = {10.1093/ehjopen/oead113},
    url = {https://doi.org/10.1093/ehjopen/oead113},
    eprint = {https://academic.oup.com/ehjopen/article-pdf/3/6/oead113/54703360/oead113.pdf},
}

@Article{Wang:2020,
author={Wang, W.
and Wang, H.
and Chen, Q.
and Zhou, Z.
and Wang, R.
and Zhang, N.
and Chen, Y.
and Sun, Z.
and Xu, L.},
title={Coronary artery calcium score quantification using a deep-learning algorithm},
journal={Clinical Radiology},
year={2020},
month={Mar},
day={01},
publisher={Elsevier},
volume={75},
number={3},
pages={237.e11-237.e16},
issn={0009-9260},
doi={10.1016/j.crad.2019.10.012},
url={https://doi.org/10.1016/j.crad.2019.10.012}
}

@article{Winkel:2021,
    author = {Winkel, David J and Suryanarayana, V Reddappagari and Ali, A Mohamed and Görich, Johannes and Buß, Sebastian Johannes and Mendoza, Axel and Schwemmer, Chris and Sharma, Puneet and Schoepf, U Joseph and Rapaka, Saikiran},
    title = {Deep learning for vessel-specific coronary artery calcium scoring: validation on a multi-centre dataset},
    journal = {European Heart Journal - Cardiovascular Imaging},
    volume = {23},
    number = {6},
    pages = {846-854},
    year = {2021},
    month = {07},
    issn = {2047-2404},
    doi = {10.1093/ehjci/jeab119},
    url = {https://doi.org/10.1093/ehjci/jeab119},
}

@article{Xu:2021,
title = {Performance of artificial intelligence-based coronary artery calcium scoring in non-gated chest CT},
journal = {European Journal of Radiology},
volume = {145},
pages = {110034},
year = {2021},
issn = {0720-048X},
doi = {https://doi.org/10.1016/j.ejrad.2021.110034},
url = {https://www.sciencedirect.com/science/article/pii/S0720048X21005155},
author = {Jie Xu and Jia Liu and Ning Guo and Linli Chen and Weixiang Song and Dajing Guo and Yu Zhang and Zheng Fang},
}

@Article{Zeleznik:2021,
author={Zeleznik, Roman
and Foldyna, Borek
and Eslami, Parastou
and Weiss, Jakob
and Alexander, Ivanov
and Taron, Jana
and Parmar, Chintan
and Alvi, Raza M.
and Banerji, Dahlia
and Uno, Mio
and Kikuchi, Yasuka
and Karady, Julia
and Zhang, Lili
and Scholtz, Jan-Erik
and Mayrhofer, Thomas
and Lyass, Asya
and Mahoney, Taylor F.
and Massaro, Joseph M.
and Vasan, Ramachandran S.
and Douglas, Pamela S.
and Hoffmann, Udo
and Lu, Michael T.
and Aerts, Hugo J. W. L.},
title={Deep convolutional neural networks to predict cardiovascular risk from computed tomography},
journal={Nature Communications},
year={2021},
month={Jan},
day={29},
volume={12},
number={1},
pages={715},
issn={2041-1723},
doi={10.1038/s41467-021-20966-2},
url={https://doi.org/10.1038/s41467-021-20966-2}
}

@inbook{PyTorch, 
author = {Paszke, Adam and Gross, Sam and Massa, Francisco and Lerer, Adam and Bradbury, James and Chanan, Gregory and Killeen, Trevor and Lin, Zeming and Gimelshein, Natalia and Antiga, Luca and Desmaison, Alban and K\"{o}pf, Andreas and Yang, Edward and DeVito, Zach and Raison, Martin and Tejani, Alykhan and Chilamkurthy, Sasank and Steiner, Benoit and Fang, Lu and Bai, Junjie and Chintala, Soumith},
title = {PyTorch: an imperative style, high-performance deep learning library},
year = {2019},
publisher = {Curran Associates Inc.}, 
address = {Red Hook, NY, USA}, 
booktitle = {Proceedings of the 33rd International Conference on Neural Information Processing Systems},
articleno = {721}, 
numpages = {12} }

@InProceedings{Adam,
  author    = {Kingma, Diederik and Ba, Jimmy},
  booktitle = {International Conference on Learning Representations (ICLR)},
  title     = {Adam: A Method for Stochastic Optimization},
  year      = {2015},
  address   = {San Diega, CA, USA},
  optmonth  = {12},
}

@ONLINE{AIHW_CVD,
    author    = "Australian Institute of Health and Welfare",
    title     = "Over a third of all deaths caused by cardiovascular disease, diabetes, and chronic kidney disease",
    publisher = "Australian Government",
    month     = "Oct",
    year      = "2014",
    url       = "https://www.aihw.gov.au/news-media/media-releases/2014/october/over-a-third-of-all-deaths-caused-by-cardiovascula",
    urldate   = "2025-10-20"
}

@ONLINE{AIHW_Heart_Stroke,
    author    = "Australian Institute of Health and Welfare",
    title     = "Heart, stroke and vascular disease: Australian facts",
    publisher = "Australian Government",
    month     = "Dec",
    year      = "2024",
    url       = "https://www.aihw.gov.au/reports/heart-stroke-vascular-diseases/hsvd-facts/contents/all-heart-stroke-and-vascular-disease/coronary-heart-disease",
    urldate   = "2025-10-20"
}

\end{document}